\documentclass[conference]{IEEEtran}
\usepackage{times}

\IEEEoverridecommandlockouts %

\usepackage[numbers]{natbib}
\usepackage{multicol}
\usepackage[bookmarks=true]{hyperref}
\usepackage{graphics} %
\usepackage{epsfig} %
\usepackage{times} %
\usepackage{amsmath} %
\usepackage{amssymb}  %
\usepackage{hyperref} %
\usepackage{xcolor} %
\usepackage{arydshln} %
\usepackage{color} 
\usepackage{float}
\usepackage{multirow}
\usepackage{booktabs}
\usepackage{tikz}
\usepackage{wrapfig}
\usepackage{units}
\usepackage{bm}
\usepackage{balance}
\usepackage{tabularx}
\usetikzlibrary{calc ,math, positioning, intersections, decorations, external, plotmarks, shapes}
\usepackage{pgfplots}
\usepackage{pgfplotstable}
\usepgfplotslibrary{statistics}
\pgfplotsset{compat=1.16}
\usepackage[skip=0pt,font=footnotesize]{caption}
\usepackage{subcaption}

\newcommand{\camera}[1]{{\color{black} #1}}

\newcommand{\wfr}[0]{\ensuremath{\mathcal{W}}} %
\newcommand{\bfr}[0]{\ensuremath{\mathcal{B}}} %
\newcommand{\cfr}[0]{\ensuremath{\mathcal{C}}} %
\newcommand{\ifr}[0]{\ensuremath{\mathcal{I}}} %

\newcommand\norm[1]{\left\lVert#1\right\rVert}

\newcolumntype{C}{>{\centering\arraybackslash}X}
\newcolumntype{P}[1]{>{\centering\arraybackslash}p{#1}}

\definecolor{matlab1}{rgb}{0.00000,0.44700,0.74100}%
\definecolor{matlab2}{rgb}{0.85000,0.32500,0.09800}%
\definecolor{matlab3}{rgb}{0.92900,0.69400,0.12500}%
\definecolor{matlab4}{rgb}{0.49400,0.18400,0.55600}%
\definecolor{matlab5}{rgb}{0.4660, 0.6740, 0.1880}%
\definecolor{matlab6}{rgb}{0.3010, 0.7450, 0.9330}%
\definecolor{matlab7}{rgb}{1.0, 0.0, 0.0}%

\def\method{HDVIO}
\def\ourMethod{HDVIO (Ours)}

\definecolor{somegray}{rgb}{0.5, 0.5, 0.5}
\newcommand{\darkgrayed}[1]{\textcolor{somegray}{#1}}
\makeatletter
\newcommand*\titleheader[1]{\gdef\@titleheader{#1}}
\AtBeginDocument{%
  \let\st@red@title\@title
  \def\@title{%
    \vskip-2em
    \bgroup\normalfont\large\centering\@titleheader\par\egroup
    \vskip0.5em\st@red@title}
}
\makeatother

\titleheader{\darkgrayed{This paper has been accepted for publication at the
Robotics: Science and Systems conference (RSS), Daegu, 2023.}}

\title{HDVIO: Improving Localization and Disturbance Estimation with Hybrid Dynamics VIO}

\begin{document}

\author{Giovanni Cioffi\authorrefmark{1}, Leonard Bauersfeld\authorrefmark{1}, Davide Scaramuzza
\thanks{\authorrefmark{1}Equal contribution.\newline
The authors are with the Robotics and Perception Group, Department of Informatics, University of Zurich, Switzerland, \protect\url{http://rpg.ifi.uzh.ch}.\newline
This work was supported by the National Centre of Competence in Research (NCCR) Robotics through the Swiss National Science Foundation (SNSF) and the European Union’s Horizon 2020 Research and Innovation Programme under grant agreement No. 871479 (AERIAL-CORE) and the European Research Council (ERC) under grant agreement No. 864042 (AGILEFLIGHT).}
}

\maketitle

\begin{abstract}
\camera{Visual-inertial odometry (VIO) is the most common approach for estimating the state of autonomous micro aerial vehicles using only onboard sensors}. 
Existing methods improve VIO performance by including a dynamics model in the estimation pipeline. 
However, such methods degrade in the presence of low-fidelity vehicle models and continuous external disturbances, such as wind. Our proposed method, \method, overcomes these limitations by using a hybrid dynamics model that combines a point-mass vehicle model with a learning-based component that captures complex aerodynamic effects. \method\ estimates the external force and the full robot state by leveraging the discrepancy between the actual motion and the predicted motion of the hybrid dynamics model. 
Our hybrid dynamics model uses a history of thrust and IMU measurements to predict the vehicle dynamics.
To demonstrate the performance of our method, we present results on both public and novel drone dynamics datasets and show real-world experiments of a quadrotor flying in strong winds up to \unit[25]{km/h}. The results show that our approach improves the motion and external force estimation compared to the state-of-the-art by up to 33\% and 40\%, respectively. 
Furthermore, differently from existing methods, we show that it is possible to predict the vehicle dynamics accurately while having no explicit knowledge of its full state.
\end{abstract}

\IEEEpeerreviewmaketitle

\section*{Supplementary Material}\label{sec:SupplementaryMaterial}

A narrated video illustrating our approach is available at: \url{https://youtu.be/CrnINdJS3s4}
\section{Introduction}\label{sec:Introduction}

Visual-inertial odometry (VIO) has become the de-facto standard for state estimation of consumer and inspection drones.
To improve the performance of the VIO pipeline, multiple approaches that tightly couple the drone dynamics in VIO systems have been recently proposed~\cite{nisar2019vimo, ding2021vid, chen2022visual}.
Including the system dynamics in the VIO formulation brings in new information, which allows the VIO system to distinguish between motion due to actuation and motion due to perturbations (external forces). 
This results in an increased accuracy of the pose estimates and the possibility to estimate an external force acting on the robot.

Despite working well in many situations, the performance of state-of-the-art methods degrades drastically if the model mismatch is large (high speeds, systematic noise in the actuation inputs) or if continuous external disturbances are present (continuous wind).
This is because their simplifying assumptions\textemdash no aerodynamic drag and zero-mean noise in the system dynamics\textemdash no longer hold.
Simply reusing a state-of-the-art dynamics model~\cite{bauersfeld2021neurobem, sun2019quadrotor} inside a VIO pipeline is difficult as the dynamics model should only depend on measurements and not rely on the full robot state, otherwise, a feedback loop is introduced where the VIO output influences the system model which then, in turn, affects the VIO.

Addressing these limitations enables the deployment of model-based VIO estimators in applications where aerodynamic effects are significant, such as during fast flights~\cite{bauersfeld2022rangeestimates} and in windy conditions~\cite{connell2022neuralfly}, or in cases where modeling inaccuracies are present.
\begin{figure}
    \centering
    \includegraphics{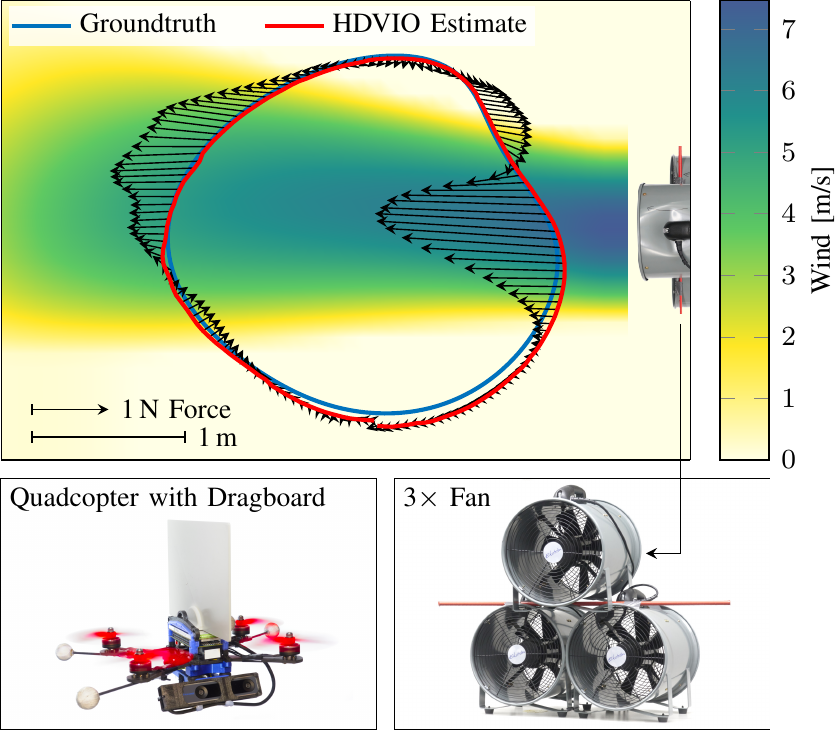}
    \vspace*{6pt}
    \caption{A quadrotor with a dragboard attached is flown on a circular trajectory through a wind field generated with three industrial fans. Our \method\ is used to estimate the position of the drone (shown in red) and the external disturbance force (black arrows) acting on the vehicle. The ground-truth position of the vehicle is shown in blue.}
    \vspace*{-12pt}
    \label{fig:fig1}
\end{figure}

The state-of-the-art approach VIMO~\cite{nisar2019vimo} integrates the drone dynamics in an optimization-based VIO~\cite{qin2018vins} system by defining new a residual term derived from the propagation of the drone model.
The drone dynamics are based on a point-mass model that neglects aerodynamic effects, leading to the aerodynamic drag being estimated as part of the external force.
Another limitation of VIMO is that any systematic offset in the actuation inputs (vehicle miscalibration, such as wrong rotor lift coefficients) is estimated as an accelerometer bias. 
This leads to erroneous information being introduced into the inertial residuals, resulting in reduced motion estimation accuracy.

All the above limitations can be addressed by improving the dynamics model of the drone included in the VIO estimator. 
Current methods to estimate high-fidelity drone models~\cite{bauersfeld2021neurobem, furrer2016rotors, gill2017propeller} typically require knowledge of the full drone state, including velocity and attitude.
However, the full state is part of the VIO output.
Consequently, the prediction of the drag force would depend on the estimator state, introducing a feedback loop that can lead to a diverging estimator.

\subsection*{Contribution}
We present \method, the first VIO pipeline, which uses a neural network to refine the drone dynamics model. 
In contrast to prior work focusing on drone modeling~\cite{bauersfeld2021neurobem}, our learned dynamics model does not require knowledge of the full drone state (for example, velocity). 
Instead, by using a temporal convolutional network~\cite{oord2016wavenet}, it only needs thrust (commands or measurements) and angular velocities from a gyroscope to estimate the aerodynamic forces.
We integrate our hybrid drone model into an optimization-based VIO system~\cite{Forster17troSVO, leutenegger2015keyframe}, and leverage the preintegration theory~\cite{forster2016manifold} to efficiently compute dynamics residuals between consecutive camera frames.
The dynamics residuals are optimized together with monocular camera and IMU residuals in the VIO backend.

We evaluate our method against the same VIO system without the proposed hybrid dynamics model and also against VIMO~\cite{nisar2019vimo} in multiple experiments using the public real-world datasets \emph{Blackbird}~\cite{antonini2020blackbird} and \emph{VID}~\cite{zhang2022visual}.
The results show that our method overcomes the limitations of current state-of-the-art methods and estimates the robot states and the external force more accurately, achieving up to $33\%$ and $40\%$ improvement, respectively.
Furthermore, we evaluate the performance of \method\ in a set of experiments where the drone is flown in a known wind field as shown in Fig.~\ref{fig:fig1}. Our method is able to outperform VIMO in the prediction of the external force due to wind. 

We also evaluate the accuracy of the learned dynamics model on the \emph{NeuroBEM}~\cite{bauersfeld2021neurobem} dataset, where it provides competitive performance compared to state-of-the-art aerodynamics models. This, for the first time, shows that the forces acting on the vehicle can be predicted without access to its full state, i.e., without knowing its ground-truth linear and angular velocity. Furthermore, to the best of our knowledge, our learning-based model is the first data-driven dynamics model that requires no ground-truth force measurements for training but only position and velocity supervision signals. We show that this removes the need for a motion-capture system to record training data, as simultaneous localization and mapping (SLAM) methods are sufficiently accurate to estimate the position and velocity ground truth for training.

By having both a precise VIO pipeline and an accurate estimate of the external force, we believe that this work is a stepping stone towards the use of \emph{autonomous} drones in safety-critical applications like surveying a disaster site and public air transport\footnote{\url{https://www.volocopter.com/newsroom/first-crewed-evtol-flight/}} which, to date, still require a human pilot.

\section{Related Work}\label{sec:RelatedWork}
Works on visual-inertial odometry with external force estimation can be separated into two groups: loosely-coupled and tightly-coupled methods. In the first category, the external force estimation for flying robots is decoupled from the motion estimation~\cite{tomic2014unified, yuksel2014nonlinear, ruggiero2014impedance, mckinnon2016unscented, augugliaro2013admittance, tagliabue2017collaborative}, whereas tightly-coupled approaches propose to simultaneously estimate the motion of the robot and the external perturbance.

\subsection{Loosely-Coupled Methods}
Initially, all developed methods~\cite{tomic2014unified, yuksel2014nonlinear, ruggiero2014impedance} were based on deterministic approaches.
They propose nonlinear force and torque observers derived from the robot dynamics model and assume that the estimates of the robot state are available from another estimator.

Differently from these methods, the probabilistic approaches proposed in~\cite{mckinnon2016unscented, augugliaro2013admittance, tagliabue2017collaborative, abeywardena2014model} account for the sensor noise and, consequently, achieve increased accuracy.
They are based on the Extended Kalman Filter (EKF)~\cite{augugliaro2013admittance, abeywardena2014model} and the Unscented Kalman Filter (UKF)~\cite{mckinnon2016unscented,tagliabue2017collaborative}.
The work in~\cite{chen2022visual} uses the quadrotor model to update an EKF-based VIO estimator~\cite{mourikis2007multi} in order to perform online system identification as well as state estimation. 
They show that in the case of noisy dynamics measurements, the best solution is to decouple the estimation of the state variables from the measurement update based on the quadrotor dynamics.
The approach in~\cite{tagliabue2020touch} uses a UKF to estimate external disturbances such as wind and interactions with humans. 
The filter is updated with the output of a neural network that processes airflow measurements from a bio-inspired airflow sensor and pose measurements from a motion capture system.

As loosely-coupled approaches neglect the correlations among the estimated variables and their noise characteristic, they suffer from a decreased performance unless the signal-to-noise ratio of the sensor data is very high.

\subsection{Tightly-Coupled Methods}
To overcome the limitations of loosely-coupled methods, more recently~\cite{nisar2019vimo, ding2021vid, fourmy2021contact} propose tightly-coupled approaches to simultaneously estimate the motion of the robot and the external perturbance.
VIMO~\cite{nisar2019vimo} is the first work that tightly couples the robot dynamics in an optimization-based VIO system~\cite{qin2018vins}.
The main contribution is the addition of a residual term that represents a motion constraint based on the robot dynamics, including external forces, to the VIO problem formulation.
The derivation of this dynamic residual term is inspired by the IMU preintegration theory~\cite{forster2016manifold}.
In this case, high-rate thrust inputs are pre-integrated, resulting in residual terms between consecutive camera frames.
The external force is modeled as a zero-mean Gaussian variable since its dynamics are unknown.
In this way, VIMO jointly estimates the external force in addition to the robot state.
However as discussed in Sec.~\ref{sec:Introduction}, this method is subject to limitations when the external force is continuous or there is a model mismatch.

The more recent method, VID-Fusion~\cite{ding2021vid}, proposes an algorithm very similar to VIMO where only the model of the external force is different.
In VID-Fusion the mean of the Gaussian distribution used to represent the external force is equal to the average difference between the accelerometer and thrust measurements in the preintegration window.
An extension of VIMO for legged robots is proposed in~\cite{fourmy2021contact}.

\subsection{Drone Modeling}
At the core of our \method\ approach, we need an accurate model of the drone dynamics. Prior work exclusively addresses this in a setting where the vehicle state is available. In a VIO pipeline, this would introduce a feedback loop inside the estimator and such a dynamics model is not suitable. Nevertheless, a brief review of quadrotor modeling literature is presented for completeness.

Often, quadrotors are modeled as a simple rigid body with mass and inertia. In this model, the robot can only exert a force in the body-z direction and has either no aerodynamic drag or linear drag \cite{furrer2016rotors, song2020flightmare, shah2018airsim, meyer2012comprehensive}. Such basic models can be refined based on first-principles, leading to blade-element-momentum (BEM) theory \cite{bauersfeld2021neurobem, gill2017propeller, hoffmann2007quadrotor, orsag2012influence}. On the other hand, purely data-driven models have been developed \cite{sun2019quadrotor} which typically outperform first-principle based methods as quadrotor aerodynamics are highly complex. The state-of-the-art model, NeuroBEM~\cite{bauersfeld2021neurobem}, combines a physical model and a learning-based component. The success of such a method has inspired us to use a learned component in our \method\ for the drone dynamics.
\section{Methodology}\label{sec:Methodology}

In this section, we describe our visual-inertial-hybrid drone dynamics odometry algorithm.
First, we introduce the notation used throughout the paper and the drone dynamics. 
\camera{We focus our derivation on a quadrotor platform, however, our approach could be extended to any other robotic platform}.
Second, we formulate the estimation problem.
Third, we present a concise derivation of the dynamics residual term. 
This derivation is based on the preintegration theory~\cite{forster2016manifold} and is also used in VIMO. 
Last, we present our learning-based module.

\subsection{Notation \& Quadrotor Dynamics}
Throughout this paper, scalars are denoted in non-bold~$[s, S]$, vectors in lowercase bold~$\bm{v}$, and matrices in uppercase bold~$\bm{M}$.
World $\wfr$, Body $\bfr$, IMU $\ifr$, and camera $\cfr$ frames are defined with an orthonormal basis, i.e. $\{\bm{x}^\wfr, \bm{y}^\wfr, \bm{z}^\wfr\}$.
The frame $\bfr$ is located at the center of mass of the quadrotor. For simplicity, the IMU frame $\ifr$ is assumed to be the same as $\bfr$.

We use the notation $(\cdot)^{\wfr}$ to represent a quantity in the world frame. 
A similar notation is utilized for every reference frame. 
The position, orientation, and velocity of $\bfr$ with respect to $\wfr$ at time $t_k$ are written as $\bm{p}_{\bfr_{k}}^{\wfr} \in \mathbb{R}^3$, $\bm{R}_{\bfr_{k}}^{\wfr} \in \mathbb{R}^{3 \times 3}$ part of the rotation group $SO(3)$, and $\bm{v}_{\bfr_{k}}^{w} \in \mathbb{R}^3$, respectively.
The unit quaternion represenation of $\bm{R}_{\bfr_{k}}^{\wfr}$ is written as $\bm{q}_{\bfr_{k}}^{\wfr}$.
The accelerometer and gyroscope bias are written as $\bm{b}_{a}$ and $\bm{b}_{g}$, respectively.
The gravity vector in the world frame is written as $\bm{g}^w$.
We use the symbol $\hat{\cdot}$ to indicate noisy measurements.

The quadrotor is assumed to be a point of mass $m$.
The evolution of the position and velocity of the quadrotor platform is described by the following model:
\begin{equation}\label{eq:drone_model}
    \dot{\bm{p}}^{\wfr}_{\bfr_k} = \bm{v}^{\wfr}_{\bfr_k},\;\;
    \dot{\bm{v}}^{\wfr}_{\bfr_k} = \bm{R}^{\wfr}_{\bfr_k} (\bm{f}^{\bfr}_{t_k} + \bm{f}^{\bfr}_{res_k} + \bm{f}^{\bfr}_{e_k}) + \bm{g}^{w},
\end{equation}
where $\bm{f}^{\bfr}_{t_k} = [0, 0, T_k]^{\top}$ is the mass-normalized collective thrust and $\bm{f}^{\bfr}_{e_k}$ is the external force acting on the quadrotor platform.
To account for aerodynamic effects and unknown systematic noise in the thrust inputs, we introduce a residual term~$\bm{f}^{\bfr}_{res_k}$.
We will drop the term mass-normalized when referring to the collective thrust hereafter for the sake of conciseness. 
The external force is assumed to be a random variable distributed according to a zero-mean Gaussian distribution. 
As pointed out in~\cite{nisar2019vimo}, this allows the estimator to
distinguish between the slowly changing accelerometer bias and the
incidental external forces.
The dynamics motion constraint is derived by leveraging the preintegration theory~\cite{forster2016manifold}.
This requires the separation of the residual terms dependent on optimization variables
from the terms dependent on the measurements. 
The rotational dynamics of the quadrotor are not considered here because the torque inputs cannot be separated from their dependency on the robot orientation.
Instead, the evolution of the orientation of the quadrotor is obtained from the gyroscope rotation model:
\begin{equation}\label{eq:gyro_model}
    \dot{\bm{q}}^{\wfr}_{\bfr_k} = \frac{1}{2} \bm{q}^{\wfr}_{\bfr_k} \otimes [0, \bm{\omega}^{\bfr_k}]^{\top},
\end{equation}
where $\otimes$ is the quaternion product and $\hat{\bm{\omega}}^{\bfr_{k}} = \bm{\omega}^{\bfr_{k}} + \bm{b}_{\omega_{k}} + \bm{n}_{\omega}$ is the gyroscope measurement.
The gyroscope noise is modeled as additive Gaussian noise $\bm{n}_{\omega} \thicksim \mathcal{N}(0, \bm{\sigma}_{\omega}^2)$. The bias is modeled as a random walk $\bm{\dot{b}}_{\omega_{k}} = \bm{n}_{b_{\omega}}$, with $\bm{n}_{b_{\omega}} \thicksim \mathcal{N}(0, \bm{\sigma}_{b_{\omega}}^2)$.

\subsection{Estimation Problem Formulation}
\begin{figure}
    \centering
    \includegraphics{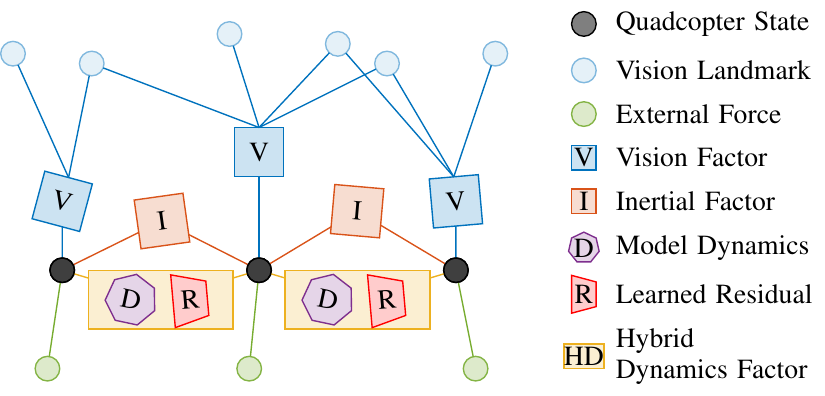}
    \vspace*{6pt}
    \caption{Factor graph representation of \method\ with visual, inertial, and hybrid dynamics factors.}
    \label{fig:factor_graph}
    \vspace*{-18pt}
\end{figure}

We implement our hybrid drone dynamics in a sliding-window optimization-based VIO system based on the nonlinear optimization proposed in~\cite{leutenegger2015keyframe}.
An overview of the proposed optimization-based VIO with hybrid drone dynamics using a factor graph representation is in Fig.~\ref{fig:factor_graph}.
The sliding window contains the most recent $L$ keyframes and $K$ drone states.
In this work, we use $L = 10$ and $K = 5$.
The optimization variables are: $\mathcal{X} = \{ \mathcal{L}, \mathcal{X}_\mathcal{L}, \mathcal{X}_{\bfr} \}$, where $\mathcal{L}$ comprises the position of the 3D landmarks seen in the sliding window, $\mathcal{X}_\mathcal{L}$ the poses of the keyframes: $\mathcal{X}_\mathcal{L} = [\bm{\zeta}_1, \cdots, \bm{\zeta}_l], \space l \in [1,L]$, and $\mathcal{X}_{\bfr}$ the poses of the drone: $\mathcal{X}_{\bfr} = [\bm{x}_1, \cdots, \bm{x}_k], \space k \in [1,K]$.
The $l^\text{th}$ keyframe pose is $\bm{\zeta}_l = [\mathbf{p}_{\bfr_l}^{\wfr}, \mathbf{q}_{\bfr_l}^{\wfr}]$. The $k^\text{th}$ drone state is $\bm{x}_{k} = [\bm{p}_{\bfr_k}^{\wfr}, \bm{q}_{\bfr_k}^{\wfr}, \bm{v}_{\bfr_k}^{\wfr}, \bm{b}_{a_k}, \bm{b}_{g_k}, \bm{f}^{\bfr}_{e_k}].$

The visual-inertial-model estimation problem is formulated as a joint nonlinear optimization that solves for the maximum a posteriori estimate of $\mathcal{X}$. The cost function to minimize is:
\begin{align}\label{eq:cost_function}
    \mathcal{L}^{\method} = & \sum_{h = 0}^{L+K-1} \sum_{j \in \mathcal{J}_h} \norm{ \bm{e}_{\bm{v}}^{j,h} }_{\bm{W}_{\bm{v}}^{j,h}}^{2} + \sum_{k = 0}^{K-1} \norm{ \bm{e}_{i}^{k} }_{\bm{W}_{i}^{k}}^{2} + \nonumber \\ & \sum_{k = 0}^{K-1} \norm{ \bm{e}_{d}^{k} }_{\bm{W}_{d}^{k}}^{2} + \norm{ \bm{e}_{m} }^{2}.
\end{align}
The cost function in Eq.~\ref{eq:cost_function} contains the weighted visual $\bm{e}_{\bm{v}}$, inertial $\bm{e}_{i}$, dynamics $\bm{e}_{d}$, and marginalization residuals $\mathbf{e}_{m}$.
The visual residuals are formulated as $\bm{e}_{\bm{v}}^{j,h} = \mathbf{z}^{j,h} - h(\mathbf{l}_{j}^{\wfr})$,
which describes the re-projection error of the landmark $\mathbf{l}_{j}^{\wfr} \in \mathcal{J}_h$, where $\mathcal{J}_h$ is the set containing all the visible landmarks from the frame $h$. The function $h(\cdot)$ denotes the camera projection model and $\mathbf{z}^{j,h}$ the 2D image measurement. We refer to~\cite{leutenegger2015keyframe} for further details.
The inertial residuals $\bm{e}_{i}$ are formulated using the IMU preintegration algorithm~\cite{forster2016manifold}.
The dynamics residuals are presented in Sec.~\ref{sec:dynamic_residual}.
The error term $\bm{e}_{m}$ denotes the prior information obtained from marginalization. We adopt the marginalization strategy proposed in~\cite{leutenegger2015keyframe}. 

We based our implementation of the sliding-window optimization on~\cite{leutenegger2015keyframe}. We merge this VIO backend with the visual frontend proposed in~\cite{Forster17troSVO}.
The code of this VIO pipeline is available open-source\footnote{\url{https://github.com/uzh-rpg/rpg_svo_pro_open}}.
We implement the drone model of VIMO and our hybrid-dynamics model in this VIO system.

\subsection{Dynamics Residual}\label{sec:dynamic_residual}

To derive the dynamics motion constraint, we use the collective force measurement model: $\hat{\bm{f}}^{\bfr}_{k} = \bm{f}^{\bfr}_{t_k} + \bm{f}^{\bfr}_{res_k}  + \bm{n}_{f_{t}}$. 
Hence, in addition to the residual force $\bm{f}^{\bfr}_{\text{res}_k}$, we also consider a zero-mean gaussian noise $\bm{n}_{f_{t}} \thicksim \mathcal{N}(0, \bm{\sigma}_{f_{t}}^2)$ to account for uncertainty in the force direction.
Given two consecutive states at $t_k$ and $t_{k+1}$, the dynamic motion constraint is:
\begin{align}\label{eq:dynamic_motion_constraint}
    \bm{e}_{d}^{k} = \begin{bmatrix} \bm{\alpha}^{\bfr_k}_{\bfr_{k+1}} - \hat{\bm{\alpha}}^{\bfr_k}_{\bfr_{k+1}} \\ \bm{\beta}^{\bfr_k}_{\bfr_{k+1}} - \hat{\bm{\beta}}^{\bfr_k}_{\bfr_{k+1}} \\
    \bm{f}_{e_k}^{\bfr}\end{bmatrix},
    \bm{W}^{k}_d = \begin{bmatrix} \bm{P}^{\bfr_{k} - 1}_{\bfr_{k+1_{[0:5]}}} & \bm{0} \\ \bm{0} & w_f \bm{I} \end{bmatrix}.
\end{align}
The quantities $\bm{\alpha}^{\bfr_k}_{\bfr_{k+1}}$ and $\bm{\beta}^{\bfr_k}_{\bfr_{k+1}}$ are the position and velocity change in the time interval $[t_k, t_{k+1}]$ and are written as:
\begin{align}
    \bm{\alpha}_{\bfr_{k+1}}^{\bfr_{k}} &= \bm{R}_{\wfr}^{\bfr_k} (\bm{p}_{\bfr_{k+1}}^{\wfr} - \bm{p}_{\bfr_{k}}^{\wfr} - \bm{v}^{\wfr}_{\bfr_{k}} \Delta t_{k} - \frac{1}{2} \bm{g}^{\wfr} \Delta t_{k}^{2}) \nonumber \\
    &- \frac{1}{2} \bm{f}^{\bfr}_{e_{k}} \Delta t_{k}^2 \nonumber \\
    \bm{\beta}_{\bfr_{k+1}}^{\bfr_{k}} &= \bm{R}_{\wfr}^{\bfr_k} (\bm{v}_{\bfr_{k+1}}^{\wfr} - \bm{v}_{\bfr_{k}}^{\wfr} - \bm{g}^{\wfr} \Delta t_{k}) - \bm{f}^{\bfr}_{e_{k}} \Delta t_{k}.
\end{align}
The quantities $\hat{\bm{\alpha}}^{\bfr_k}_{\bfr_{k+1}}$ and $\hat{\bm{\beta}}^{\bfr_k}_{\bfr_{k+1}}$ are the preintegrated position and velocity in $[t_k, t_{k+1}]$.
They can be calculated in the discrete-time using Euler numerical integration over the timestep $\delta t$:
\begin{align}
    \hat{\bm{\alpha}}_{i+1}^{\bfr_{k}} & = \hat{\bm{\alpha}}_{i}^{\bfr_{k}} + \hat{\bm{\beta}}_{i}^{\bfr_{k}} \delta t + \frac{1}{2}\bm{R}(\hat{\bm{\gamma}}_{i}^{\bfr_{k}})\hat{\bm{f}^{\bfr}_{i}} \delta t^2 \nonumber \\
    \hat{\bm{\beta}}_{i+1}^{\bfr_{k}} & = \hat{\bm{\beta}}_{i}^{\bfr_{k}} + \bm{R}(\hat{\bm{\gamma}}_{i}^{\bfr_{k}})\hat{\bm{f}^{\bfr}_{i}} \delta t \nonumber \\
    \hat{\bm{\gamma}}_{i+1}^{\bfr_{k}} & = \hat{\bm{\gamma}}_{i}^{\bfr_{k}} \otimes \begin{bmatrix} 1 \\ \frac{1}{2}(\hat{\bm{\omega}}^{\bfr}_{i} - \bm{b}_{\omega_{k}})\delta t \end{bmatrix},
    \label{eq:preintegration_terms_recursive_formulation}
\end{align}
where the initial conditions are: $\hat{\bm{\alpha}}_{\bfr_{k}}^{\bfr_{k}} = \hat{\bm{\beta}}_{\bfr_{k}}^{\bfr_{k}}$ = 0 and $\hat{\bm{\gamma}}_{\bfr_{k}}^{\bfr_{k}}$ is equal to the identity quaternion. $\bm{R}(\hat{\bm{\gamma}}_{i}^{\bfr_{k}})$ is the rotation matrix representation of $\hat{\bm{\gamma}}_{i}^{\bfr_{k}}$.
We run the propagation algorithm at the rate of the IMU, which is the faster sensor in our experiments.
The quantity $\bm{W}^{k}_d$ is the weight of the residual term. 
It can be calculated from the 6$\times$6 top-left block of the covariance $\bm{P}^{\bfr_{k}}_{\bfr_{k+1}}$. This covariance is derived by linearizing the error, $\delta \bm{z} = [\delta \bm{\alpha}, \delta \bm{\beta}, \delta \bm{\theta}, \delta \bm{b}_{\omega}]^{\top}$, and noise, $\bm{n} = [\bm{n}_{f_t}, \bm{n}_{\omega}, \bm{n}_{b_{\omega}}]^{\top}$ in $\delta t$. Where $\delta \bm{\theta} = \bm{\gamma}_{\bfr_{k+1}}^{\bfr_{k}} - \hat{\bm{\gamma}}_{\bfr_{k+1}}^{\bfr_{k}}$, with $\bm{\gamma}_{\bfr_{k+1}}^{\bfr_{k}} = \bm{q}_{\wfr}^{\bfr_k} \otimes \bm{q}^{\wfr}_{\bfr_{k+1}}$, and $\delta \bm{b}_{\omega} = \bm{b}_{\omega_{k+1}} - \bm{b}_{\omega_{k}}$.
It is important to note that the preintegrated terms depend on the gyroscope bias. In order to avoid repropagating each time that the estimate of the gyroscope bias changes, we employ the strategy proposed in~\cite{forster2016manifold}. Namely, the preintegration terms are corrected by their first-order approximation with respect to the change in the gyroscope bias.
We refer to~\cite{forster2016manifold} for a detailed derivation of the preintegration theory.

\subsection{Learning Residual Dynamics}
\label{sec:LearningResidualDynamics}

The dynamics residual term presented above relies on an accurate estimate of the forces acting on the vehicle. 
In previous works, modeling aerodynamic effects, such as drag forces, requires knowledge of the linear velocity of the vehicle, which is not measured, but is part of the state to be estimated. 
Therefore, simply reusing a state-of-the-art quadcopter dynamics model is not possible.

In our setting, we have access to thrusts and gyroscope measurements as these values are directly measured. The goal is to estimate a residual force $\bm f^{\bfr}_{res}$ that accounts for aerodynamic effects and model mismatches (systematic noise) between the commanded or measured thrust $T$ and the actual force acting on the robot (when no disturbance is present).
\begin{figure}
    \centering
    \includegraphics{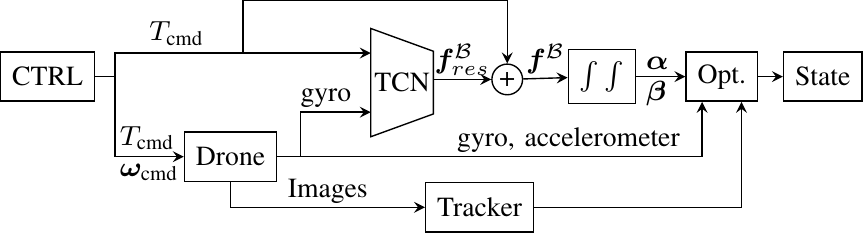}
    \vspace*{-3pt}
    \caption{Our novel learning-based model learns to predict the forces acting on the vehicle only based on the thrusts commanded by the controller and the gyro measurements from the IMU onboard the drone. The commanded collective thrust is added to the predicted residual force $\bm f^{\bfr}_{res}$ and then integrated to obtain the velocity and position updates. The optimizer then estimates the vehicle state in a tightly-coupled fashion, taking into account the displacements predicted by the hybrid dynamics model, the displacements predicted by the IMU, and the visual features.}
    \label{fig:flowchart}
    \vspace{-14pt}
\end{figure}

For this, we propose the system architecture shown in Fig.~\ref{fig:flowchart}. At the core of the learning-based component, we use a temporal convolutional network (TCN). TCNs have been shown to be as powerful as recurrent networks to model temporal sequences~\cite{bai2018empirical} but require less computation. The network takes a buffer of collective thrust and gyroscope measurements as input. The bias is removed from the gyroscope measurements.
During training, we assume that the bias behaves as a random Gaussian variable with zero mean and standard deviation equal to 1e-3. 
At deployment time, we use the current bias estimate.
Given as input a buffer of measurements in the time interval $\Delta t_{i,j} = t_{j} - t_i$, the neural net output is the residual force $\bm{f}^{\bfr}_{res_i}$.
This residual is added to the thrust inputs $\bm{f}^{\bfr}_{t_k}$ with $k \in [t_i, t_j]$ to yield forces $\hat{\bm{f}}^{\bfr}_{k}$ that take aerodynamics and robot miscalibration into account.
These forces are then used inside the preintegration framework, see Sec.~\ref{sec:dynamic_residual}, to obtain relative velocity and position measurements.

We train the neural network to minimize the MSE loss:
\begin{align}\label{equ:net_loss}
    \mathcal{L}^{HD}(\Delta \bm{\alpha}, & \Delta \hat{\bm{\alpha}}, \Delta \bm{\beta}, \Delta \hat{\bm{\beta}}) = \nonumber \\ & \frac{1}{N} \sum_{n = 1}^{N} (\norm{\bm{\alpha}^{\bfr_j}_{\bfr_i} - \hat{\bm{\alpha}}^{\bfr_j}_{\bfr_i}}^{2} + \norm{\bm{\beta}^{\bfr_j}_{\bfr_i} - \hat{\bm{\beta}}^{\bfr_j}_{\bfr_i}}^{2})
\end{align}
where $\bm{\alpha}^{\bfr_j}_{\bfr_i}$ and $\bm{\beta}^{\bfr_j}_{\bfr_i}$ are the ground-truth velocity and position changes, and $N$ is the batch size.
In order to learn aerodynamic effects and systematic noise in the input thrust measurements, there is no external force acting on the drone in the training data. Also, our training does not necessarily require force ground-truth data, which removes the need for a high-resolution motion-capture system. Instead, our training data could be collected using a SLAM pipeline.

\camera{Our TCN architecture consists of four temporal convolutional layers with 64 filters, followed by three temporal
convolutional layers with 128 filters each. A final linear layer
maps the signal to a 3-dimensional vector representing the learned residual thrust.}
The thrust and IMU measurements are sampled at 100 Hz and are fed to the TCN in an input buffer of a length of 100 ms.
\camera{Consequently, each input buffer contains 10 thrust and 10 gyroscope measurements.}
We use the Gaussian Error Linear Unit (GELU) activation function.
We train our neural network on a laptop running Ubuntu 20.04 and equipped with an Intel Core i9 2.3GHz CPU and Nvidia RTX 4000 GPU.
Training is performed using the Adam optimizer with an initial learning rate of 1e-4.
To test the feasibility of deploying the neural network onboard the quadrotor, we test the neural net inference
on an NVIDIA Jetson TX2, which is the computing platform onboard the quadrotor.
The neural net inference runs at $\approx$180 Hz on an NVIDIA Jetson TX2 which exceeds the required \unit[100]{Hz} state-update rate of our controllers for agile flight.
\section{Results on Datasets}\label{sec:ResultsOnDatasets}

In our experiments, we evaluate our method against the same VIO system without the proposed hybrid-dynamics model (from now on referred to as VIO) and also against VIMO.
We refer the reader to the Appendix for an evaluation against VID-Fusion~\cite{ding2021vid}.
Following the best practices in the evaluation of VIO algorithms~\cite{Zhang18iros}, we use the evaluation metrics: translation absolute trajectory error ($\text{ATE}_{\text{T}}$ [m]), rotation absolute trajectory error ($\text{ATE}_{\text{R}}$ [deg]), and relative translation and rotation errors. 
These error metrics are computed after aligning the estimated trajectory with the pose-yaw method~\cite{Zhang18iros}.
We refer the reader to~\cite{Zhang18iros} for a detailed description of these metrics.
In addition, we use the RMSE between the ground truth and predicted force to evaluate the accuracy of force estimation.

\subsection{NeuroBEM Dataset}\label{sec:NeuroBEMDataset}
\subsubsection*{Experimental Setup}
In the first set of experiments, we want to evaluate the hybrid dynamics model without the full VIO pipeline. That is, the estimate of the total force $\bm f^{\bfr}$ (see Fig.~\ref{fig:flowchart}) acting on the quadcopter from a history of thrusts and gyroscope measurements and compare it to ground-truth data. To perform this evaluation, we use the challenging NeuroBEM~\cite{bauersfeld2021neurobem} dataset. It features data from indoor drone flights at speeds up to \unit[65]{km/h}. 
This dataset provides rotor speeds, from which we compute thrust measurements, and gyroscope measurements alongside ground-truth forces.

We compare the force estimation accuracy of our learned dynamics model to the state-of-the-art baselines. 
The Quadratic Fit is the model used in VIMO. 
BEM is a state-of-the-art first-principles model which models the force and torque acting on a propeller by integrating over infinitesimal area elements of the propeller \cite{gill2017propeller, bauersfeld2021neurobem}. 
The PolyFit model \cite{sun2019quadrotor} is a data-driven model which relies on polynomial basis functions to model the drone dynamics. 
Finally, NeuroBEM is a hybrid model which augments the BEM model with a learning based component. 
Note that all baselines except the Quadratic Fit model require the full vehicle state as an input, including the linear and angular velocities.

\begin{table}[t]
\centering
\caption{Comparison in terms of RMSE of the force estimates on the test set of the NeurBEM dataset. Methods of Type DD are data-driven, as opposed to FP first-principles methods. Our method performs remarkably well given it has no information about the velocity or orientation of the vehicle and only falls short of the NeuroBEM method which has access to the full vehicle state. The values for the first four methods are taken from \cite{bauersfeld2021neurobem}.}
\label{tab:forces_torques}
\vspace*{4pt}
\setlength{\tabcolsep}{3pt}
\begin{tabularx}{1\linewidth}{lll|cC|C}
\toprule
Model & Type & Inputs  & $F_\text{xy}$ [N]     & $F_\text{z}$ [N]     & $F$ [N] \\
\midrule
Quadratic Fit & FP & thrust & 1.536   & 1.381   & 1.486  \\
BEM \cite{bauersfeld2021neurobem} & FP & full state & 0.803   & 1.265 & 0.982 \\
PolyFit~\cite{sun2019quadrotor} & DD & full state & 0.453   & 0.832 & 0.606  \\
NeuroBEM \cite{bauersfeld2021neurobem} & DD &  full state & \textbf{0.204} & \textbf{0.504}  & \textbf{0.335} \\
\midrule
\ourMethod & DD & thrust + gyro & 0.402 & 0.672 & 0.491  \\ 
\bottomrule
\end{tabularx}
\end{table}
\begin{figure}
    \centering
    \includegraphics{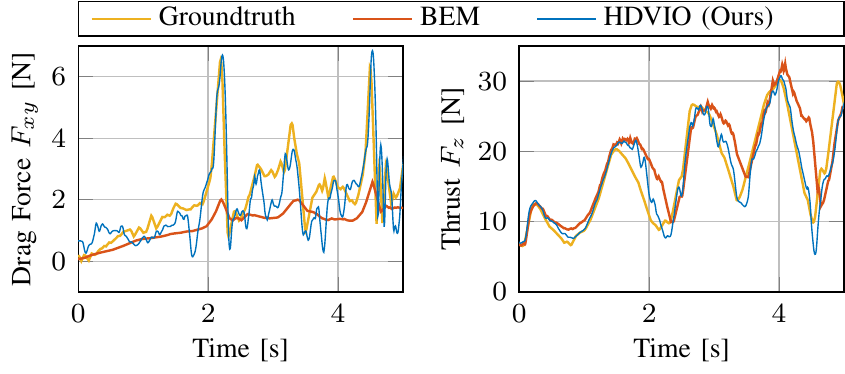}
    \caption{This figure illustrates the results shown in Tab.~\ref{tab:forces_torques} by exemplarily showing the force estimates on a very fast trajectory from the NeuroBEM dataset. Our \method\ clearly outperforms the state-of-the-art first-principle model BEM and is able to model aerodynamic effects accurately on short timescales.}
    \label{fig:neurobem_plot}
    \vspace*{-12pt}
\end{figure}

\subsubsection*{Evaluation}
The results of the comparison are summarized in Tab.~\ref{tab:forces_torques}.
As expected, the NeuroBEM method that has access to the full robot state outperforms ours. However our \method~outperforms BEM by a factor of three and PolyFit by a factor of two. 
Figure~\ref{fig:neurobem_plot} illustrates this: it shows the ground-truth forces and the forces estimated by BEM and our \method\ during the first five seconds of a very fast flight where the vehicle accelerates to \unit[15]{m/s} on a lemniscate track. 
The accuracy of our \method\ is remarkable because, in contrast to the baselines, it has no access to the ground-truth state information like the linear or angular velocity of the vehicle.

From this experiment, we conclude that our learning-based component is able to accurately capture the aerodynamic forces acting on the quadrotor and, consequently, is suitable for integration into a VIO pipeline. 
The fact that it is possible to estimate the aerodynamic forces so accurately only through a history of thrusts and gyroscope measurements is an interesting finding by itself.

\subsection{Blackbird Dataset}\label{sec:BlackbirdDataset}
\subsubsection*{Experiment Setup} 
In this second set of experiments, we evaluate our system and the baselines on the Blackbird dataset~\cite{antonini2020blackbird}.
The Blackbird dataset provides rotor speed measurements recorded onboard a quadrotor flying in a motion-capture system. We use these measurements to compute mass-normalized collective thrust measurements.
In addition, this dataset also contains IMU measurements and, in some of the sequences, photorealistic images of synthetic scenes.
In total, the Blackbird dataset contains 18 different trajectories with speeds from \unit[0.5]{m/s} to \unit[9.0]{m/s}.
Since this dataset does not contain external disturbances, we only evaluate the accuracy of the pose estimates.
Among the trajectories with available camera images, we select 6 for evaluation: \textit{Bent Dice}, \textit{Clover}, \textit{Egg}, \textit{Mouse}, \textit{Star}, and \textit{Winter}.
The remaining \unit[80]{\%} of the trajectories, amounting to approximately \unit[2]{hrs} of flight data, are used for training and \unit[20]{\%} for validation of our neural network.
In addition, we also train our network on a reduced training dataset that contains speeds up to \unit[2]{m/s} to evaluate the generalization performance to higher speeds than the ones present in the training data.

\begin{table}[t!]
\caption{
Evaluation of the trajectory estimates in the Blackbird dataset. In bold are the best values, and in underlined are the second-best values. \method$^*$ (ours) is trained on a reduced training set, with speeds up to \unit[2]{m/s} to evaluate generalization performance.}
\vspace{3pt}
\label{tab:blackbird}
\setlength{\tabcolsep}{4pt}
\begin{tabularx}{1.0\linewidth}{P{1.2cm}P{0.8cm}|C|C|C|C}
\toprule 
\multirow[c]{2}{=}[-6pt]{\centering Trajectory\newline Name} & \multirow[c]{2}{=}[-6pt]{\centering $v_{\max}$ \newline [\unit{m/s}]} & \multicolumn{4}{c}{Evaluation Metric: $\text{ATE}_\text{T}$ [\unit{m}] / $\text{ATE}_\text{R}$ [\unit{deg}]}  \\[4pt]
& & VIO & VIMO & \method \newline (ours) & \method$^*$ \newline (ours) \\
 \midrule
 Bent Dice & 3 & \textbf{0.20} / 1.78 & 0.31 / \underline{1.53} & \underline{0.21} / \underline{1.53} & 0.23 / \textbf{1.46} \\
 Clover & 5 & 0.90 / 3.52 & 0.88 / 3.66 & \underline{0.60} / \textbf{2.08} & \textbf{0.59} / \underline{2.95} \\
 Egg & 5 & 1.07 / 1.54 & 0.75 / 1.34 & \textbf{0.59} / \textbf{1.21} & \underline{0.68} / \underline{1.28} \\
 Egg & 6 & 1.40 / 2.35 & 0.98 / 4.89 & \underline{0.83} / \textbf{1.62} & \textbf{0.81} / \underline{2.26} \\
 Egg & 8 & 1.79 / 4.55 & 1.57 / 3.69 & \textbf{1.06} / \textbf{2.89} & \underline{1.22} / \underline{3.52} \\
 Mouse & 5 & 1.10 / 4.54 & 0.76 / 2.14 & \textbf{0.36} / \textbf{1.40} & \underline{0.40} / \underline{1.75} \\
 Star & 1 & \underline{0.17} / 0.78 & 0.18 / 1.05 & \textbf{0.16} / \textbf{0.58} & \textbf{0.16} / \underline{0.62} \\
 Star & 3 & 0.62 / 3.50 & 0.43 / \textbf{1.38} & \underline{0.38} / \underline{1.40} & \textbf{0.34} / 1.42 \\
 Winter & 4 & 0.97 / 2.92 & 0.69 / 2.46 & \underline{0.57} / \textbf{1.54} & \textbf{0.50} / \underline{2.27} \\
\bottomrule
\end{tabularx}
\end{table}

\begin{figure}[!]
    \centering
    \includegraphics{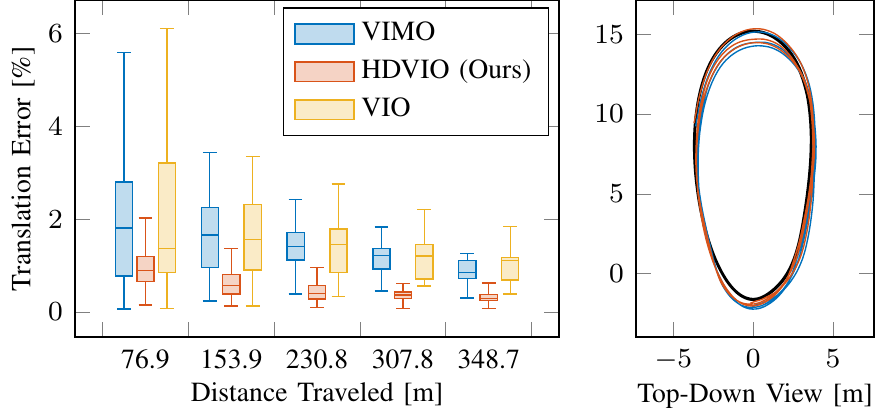}
    \caption{The performance of VIO, VIMO and \method\  is compared on the \emph{Egg} \unit[8]{m/s} trajectory (ground truth in black) from the Blackbird dataset. Our proposed method outperforms VIMO and the VIO clearly.}
    \label{fig:blackbird}
    \vspace*{-12pt}
\end{figure}

\subsubsection*{Evaluation}
We present the $\text{ATE}_{\text{T}}$ and $\text{ATE}_{\text{R}}$ on the evaluation sequences in Table~\ref{tab:blackbird}.
Our approach outperforms the VIO and VIMO baselines showing the effectiveness of our hybrid drone model.
As expected, the performance increase becomes larger at higher speeds, reaching an improvement of \unit[41]{\%} and \unit[33]{\%} against the VIO and VIMO respectively, at the max peak velocity of \unit[8]{m/s}.
This result is explained by the fact that our method includes the learned drag forces as measurements in the dynamics motion constraint.
We include the relative translation error and the top-down view of the estimated trajectory in Fig.~\ref{fig:blackbird}.
Remarkably, our system still outperforms the baselines in almost all the sequences when the network is trained on the reduced training dataset, see the last column of Tab.~\ref{tab:blackbird}.
\camera{This result shows that \method\ is able to generalize to velocities up to 4x larger than the ones included in the training data.}

\subsection{VID Dataset}\label{sec:VIDDataset}
\subsubsection*{Experiment Setup} 
In this third set of experiments, we are interested in evaluating the ability of our system in estimating an external force acting on the quadrotor and in testing our learned component when ground-truth data from a motion capture system is not available.
The VID dataset~\cite{ding2021vid} contains visual, inertial, actuation inputs, and ground-truth force measurements.
In this dataset, the data is recorded onboard a quadrotor flying in an office room equipped with a motion-capture system and in an outdoor parking area.
We use the provided rotor speed measurements to compute the thrust measurements.
We use the indoor sequences to evaluate the estimation of the external force.
Parts of these sequences include ground-truth force data.
We use the outdoor sequences to validate our learned module when ground-truth position and velocity training data is obtained from a visual-inertial SLAM system~\cite{cioffi2022continuous} instead of a motion-capture system.
Since outdoor sequences do not include ground-truth force measurements, we are interested in the estimation of the drone poses.
The quadrotor mass changes between the indoor and outdoor sequences, for this reason, we train two different neural networks, one for the indoor drone configuration and one for the outdoor drone configuration. 
We use the indoor sequences without any external perturbance to train our neural network for the indoor configuration.
They consist of a hover, a circle, and, a figure 8 trajectory amounting to only \unit[6]{min} of flying data. 
$80\%$ of this data is used for training and $20\%$ for validation.
The outdoor sequences contain circle, figure 8, and rectangle trajectories.
We use one figure 8 and one rectangle trajectory for testing.
The remaining data, amounting to only \unit[11]{min} of flying data, is used for training, $80\%$, and validation, $20\%$.
\begin{figure}
    \centering
    \includegraphics{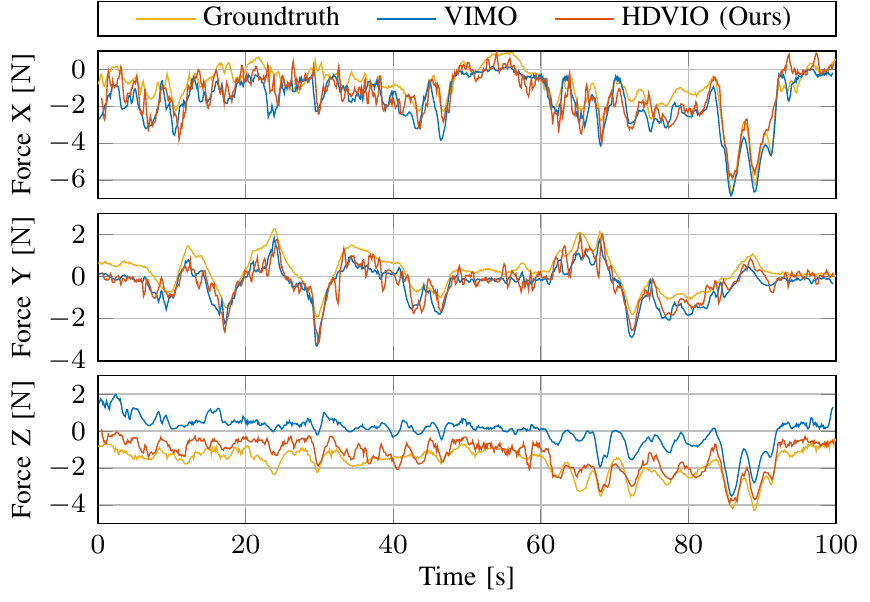}
    \caption{Comparison of the external force estimate in the \textit{sequence 17} of the VID dataset. \method\ drastically improves the force estimation along the z axes resulting in a $40\%$ reduction of the RMSE.}
    \label{fig:viddataset}
    \vspace{-9pt}
\end{figure}

\begin{figure}
    \centering
    \includegraphics{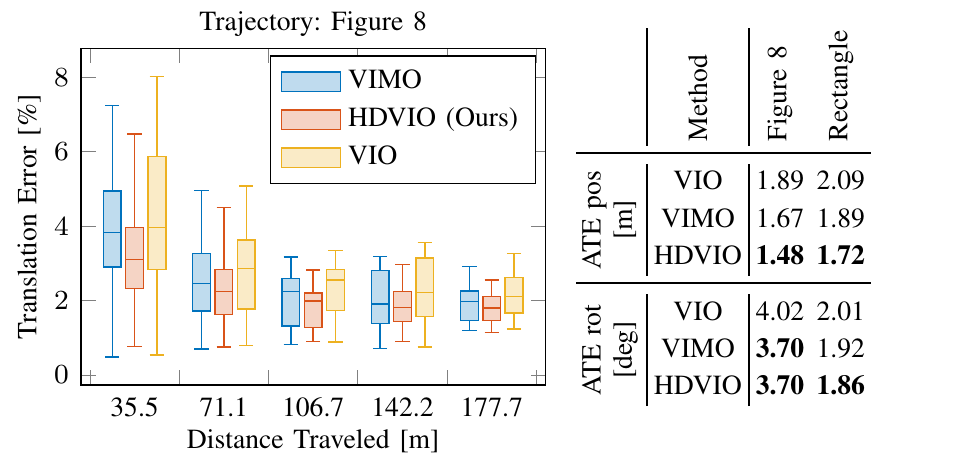}
    \caption{The plot and accompanying table show how our \method\ performs in a setting where the training data for the learning-based component is gathered from a vision-based SLAM system in the outdoor sequences of the VID dataset. The flown trajectories are at low speeds below \unit[3]{m/s}, which is why all three methods show good performance, with \method\ being the more accurate. }
    \label{fig:colmap}
    \vspace{-15pt}
\end{figure}

\subsubsection*{Evaluation}
We include in Fig.~\ref{fig:viddataset} the external force estimates of VIMO and our method in the \textit{sequence 17} of the dataset.
In this sequence, the quadrotor flies at low speed, below \unit[1]{m/s}, and is attached to a rope.
The estimates are aligned to the motion-capture reference frame.
To perform the alignment, we use the \textit{posyaw} alignment method~\cite{Zhang18iros}.
The RMSE achieved by VIMO is \unit[1.08]{N}, and the RMSE the along x, y, and z axis are 0.89, 0.63, and, \unit[1.73]{N}, respectively.
The RMSE achieved by \method\ is \unit[0.65]{N} resulting in a $40\%$ reduction.
The RMSE along the x, y, and z axis are 0.81, 0.61, and, \unit[0.55]{N}, respectively.
Clearly from the plot in Fig.~\ref{fig:viddataset}, we see that our method drastically improves the force estimation along the z axes.
Our neural network has learned to compensate for a systematic residual error affecting the thrust inputs.
We believe that the cause of this error is inaccuracy in the thrust coefficients used to compute the collective thrust inputs from the rotor speed measurements.
In this sequence, the slow motion of the vehicle and the rich texture environment renders the pose estimation problem simple.
The VIO algorithm, VIMO, and \method\ achieve similar performance.
Namely, the $\text{ATE}_{\text{T}}$ of the three methods is \unit[0.02]{m}.
The plot and accompanying table in Fig.~\ref{fig:colmap} shows the evaluation of the pose estimates in the outdoor sequences.
The flown trajectories are at low speeds below \unit[3]{m/s}, which is why all three methods show good performance, with \method\ being the more accurate.
This experiment shows that our learning-based dynamics model can be purely trained without using an external motion-capture system.
\section{Real-World Experiments}\label{sec:Setup}

In this set of experiments, we demonstrate that \method\ is able to estimate continuous external disturbances, such as continuous wind, outperforming the state-of-the-art method VIMO.
To achieve so, we fly a quadrotor in a wind field as shown in Fig.~\ref{fig:fig1}.
Details on the quadrotor platform are given in~\cite{foehn2022agilicious}.
We obtain camera and IMU measurements from an onboard Intel RealSense T265\footnote{\url{https://www.intelrealsense.com/wp-content/uploads/2019/09/Intel_RealSense_Tracking_Camera_Datasheet_Rev004_release.pdf}} camera.
Although this camera provides stereo fisheye images, we only use images from the left camera.
Rotor speed measurements are not available on our quadrotor platform.
Instead, we use the collective thrust commands that are output by the MPC controller~\cite{foehn2022agilicious} used to control the vehicle.
The quadrotor is flown in a motion-capture system that provides pose at \unit[200]{Hz} update rate.

We conduct the experiments with two different quadrotor configurations: one where the vehicle was in its nominal state and one with a {\unit[22]{cm}$\times$\unit[16]{cm}} large dragboard attached to the vehicle. The dragboard is attached such that its normal corresponds to the body y-axis of the drone. The dragboard increases the drag in the y-direction over a factor of 2 and additionally makes the vehicle much more sensitive to crosswinds.

\subsection{Wind Generation}

To generate a windfield, we place three axial fans (Ekström 12 inch, see Fig.~\ref{fig:fig1}) in an office-like room of \unit[8x10]{m} at a height of \unit[1.6]{m} above the ground, in a way that the fans are slightly angled inwards. This ensures high windspeeds across a virtual tube in front of the fans. 
Each fan has an (advertised) air circulation of $\unit[1.3]{m^3/s}$ and a measured wind speed of up to \unit[8]{m/s} at the front grill. 

To quantitatively evaluate the performance of our method, ground-truth data for the external wind forces is required.
Following, we describe how we obtain the ground-truth wind forces.

\subsubsection{Wind Speed Map}
In the first step, the wind produced by our experimental setup is measured. The local wind speed is recorded at 50 points in the wind cone in front of the fans, where samples are denser in regions where the wind speed changes quickly. The data is recorded using a hand-held anemometer (Basetech BS-10AN) whose position is tracked using the motion-capture system. To obtain the ground-truth wind speed map shown in Fig.~\ref{fig:fig1}, a smoothing spline is fitted to the data.

\begin{figure*}[t!]
    \centering
    \includegraphics{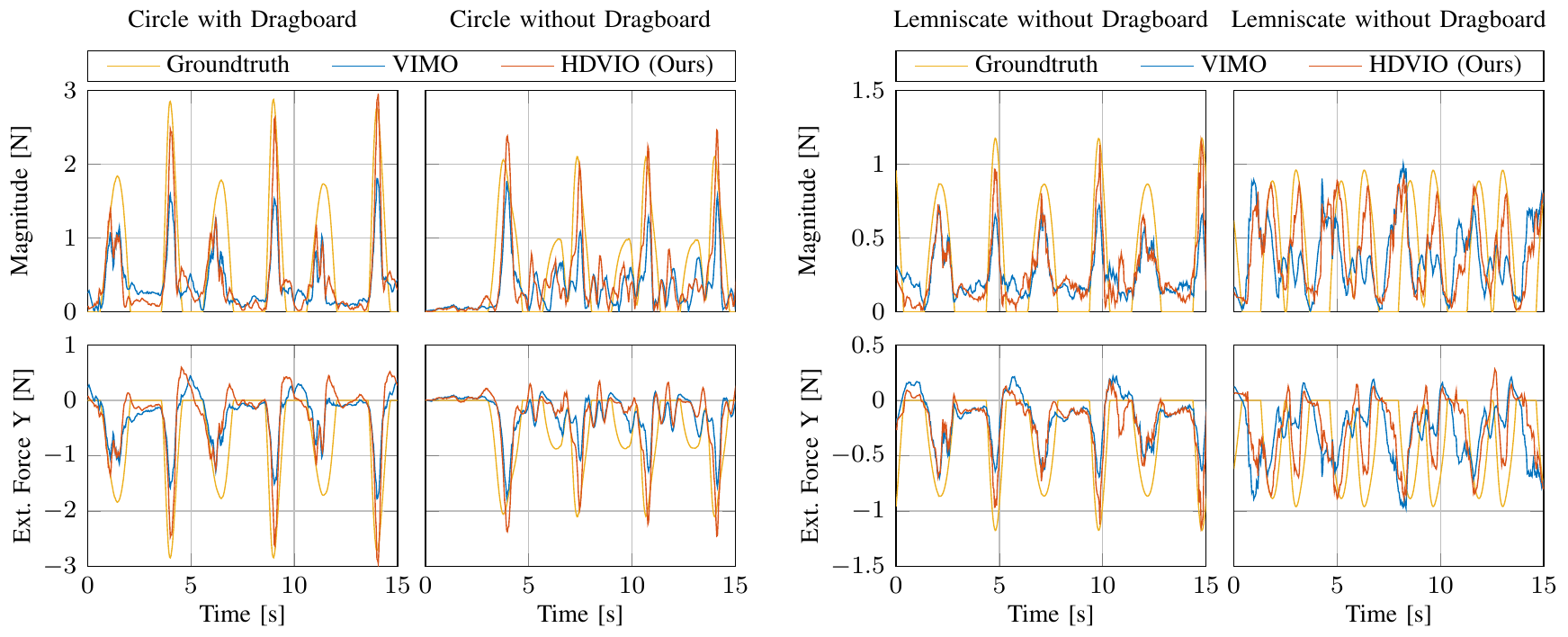}
    \caption{Wind disturbance estimates in our real-world experiments. The magnitude and the y-axis component of the wind force estimated by \method\ and VIMO. Left: circle trajectory. Right: Lemniscate trajectory. In all the plots, it is visible that \method\ achieves more accurate force estimates than VIMO.}
    \label{fig:forces_realworld}
    \vspace*{-9pt}
\end{figure*}

\subsubsection{Lift and Drag Coefficients}
Since \method\ estimates the disturbance force acting on the vehicle, we calculate the wind force based on the wind speed map. The aerodynamic forces acting on a quadrotor are primarily determined by the body/fuselage drag, $f_d^\text{fus}$, the induced drag from the propellers $f_{d}^\text{ind}$, and the lift and drag incurred by the flat-plate drag board attached to the top of the quadrotor, $f_l^\text{brd}$ and $f_d^\text{brd}$. The magnitudes of the forces can be approximated as \cite{bauersfeld2021neurobem, bauersfeld2022rangeestimates, ducard2014modeling}:
\begin{equation}
\begin{aligned}
    f_d^\text{fus} &= 0.5 \, \rho \, A^\text{fus} \, c_d^\text{fus} \, v_\text{rel}^2\; \\
    f_d^\text{ind} &= k \, v_\text{rel}\; \\
    f_{l|d}^\text{brd} &= 0.5 \, \rho \, A^\text{brd} \, c_{l|d}^\text{brd}(\alpha) \, v_\text{rel}^2\;,
\end{aligned}
\label{eq:aero_forces}
\end{equation}
where $\rho$ is the air density, $A$ is a surface area, $v_\text{rel}$ the relative air speed, $\alpha$ is the angle of attack of the dragboard, $k$ is the propeller drag coefficient, and $c_{l|d}$ are the lift and drag coefficients of the fuselage and dragboard. The relative air speed is given as the norm of the relative velocity, i.e., the sum of the ego motion and the wind.

In this model, the fuselage is represented by a square prism with an angle-of-attack independent drag coefficient of $c_d^\text{fus} = 2.0$ \cite{hinsberg2021prismaero}. For the flat-plate wing, we use a simple model for high angles-of-attack that has found widespread application in propeller modeling \cite{gill2017propeller, ducard2014modeling} and fits the  experimental data for flat-plate wings \cite{sheldahl1981aerodynamic}:
\begin{align*}
    c_l^\text{brd}(\alpha) = \sin\left( 2\alpha\right)\;, && c_d^\text{brd}(\alpha) = 2 \sin^2(\alpha).
\end{align*}
To validate the predicted forces for the fuselage and drag board in Eq.~\eqref{eq:aero_forces}, the quadrotor has been mounted on a load cell\footnote{\url{https://www.ati-ia.com/products/ft/ft_models.aspx?id=Mini40}}. At \unit[7]{m/s} wind speed, the lift and drag measurements are within \unit[10]{\%} of the calculated values. Furthermore, the linear propeller drag coefficient is found to be \unit[$k=0.145$]{Ns/m}.

\subsubsection{Wind Forces}
Our method separates aerodynamic effects, i.e., body drag and induced drag, from external forces. Therefore, to obtain the ground truth for the disturbance caused by the wind, we calculate the forces acting on the quadrotor when the fans are active and subtract the force calculated when the fans are turned off.

\begin{table}[!]
    \centering
    \setlength{\tabcolsep}{3pt}
    \caption{Trajectories estimates in our real-world experiments. We use (d) to indicate that a drag board was attached to the drone.}
    \begin{tabularx}{1\linewidth}{l|CCC|CCC}
    \toprule
         &  \multicolumn{3}{c|}{ATE Position [m]}   & \multicolumn{3}{c}{ATE Rotation [deg]} \\
         & VIO & VIMO & \method & VIO & VIMO & \method \\[3pt]
         \midrule
         Circle (d) & \bf 0.07 & 0.10 & \bf 0.07 & 2.02 & \bf 1.80 & 2.06 \\
         Circle & \bf 0.06 & 0.08 & \bf 0.06 & 1.21 & 1.19 & \bf 1.17 \\
         Lemniscate (d) & 0.38 & 0.34 & \bf 0.30 & \bf 2.39 & 2.93 & 2.81 \\
         Lemniscate & 0.27 & 0.32 & \bf 0.20 & 2.44 & 1.93 & \bf 1.84 \\
         \bottomrule
    \end{tabularx}
    \label{tab:realworld_results}
\end{table}

\subsection{Dataset Collection}\label{sec:RealworldExp}
In this set of experiments, we fly the quadrotor in a wind field with wind gusts up to \unit[25]{km/h}.
The training data consists of approximately \unit[10]{min} of random trajectories flown without wind.
We use $80\%$ of this data to train the neural network and the remaining $20\%$ for validation.
We exclusively use \emph{random} trajectories which are generated by sampling position data using a Gaussian Process. This approach ensures diverse training data and prevents overfitting to specific trajectories.
The test trajectories consist of a circle and a lemniscate trajectory with a max speed of \unit[2]{m/s}.
We also recorded a second data set which features the same training, validation, and test trajectories, but the quadrotor is equipped with a drag board. 
In this case, we want to increase the magnitude of the drag and the external force due to the wind gusts to highlight the advantage of \method\ compared to VIMO.

\subsection{Evaluation}\label{sec:RealworldExpEval}
We present the estimate of the external force due to the wind gusts in Fig.~\ref{fig:forces_realworld}.
Since the wind gusts hit the quadrotor along the y axes of the world reference frame, we show the y component of the estimated force as well as the force norm.
The external forces are estimated in the body frame of the quadrotor, cf. Eq.~\ref{eq:drone_model}.
We align them to the world frame, which corresponds to the motion-capture reference frame, using the ground-truth orientations. Using the ground-truth orientations allows us to directly compare the estimates of our method against the ones of VIMO.
From Fig.~\ref{fig:forces_realworld}, we see that our method achieves more accurate force estimates than VIMO.
In particular, \method\ outperforms VIMO in the prediction of the wind gusts when the quadrotor enters the wind field. 
This is visible from the fact that our method accurately predicts the peaks of the wind force.

\begin{figure}[!]
    \centering
    \includegraphics{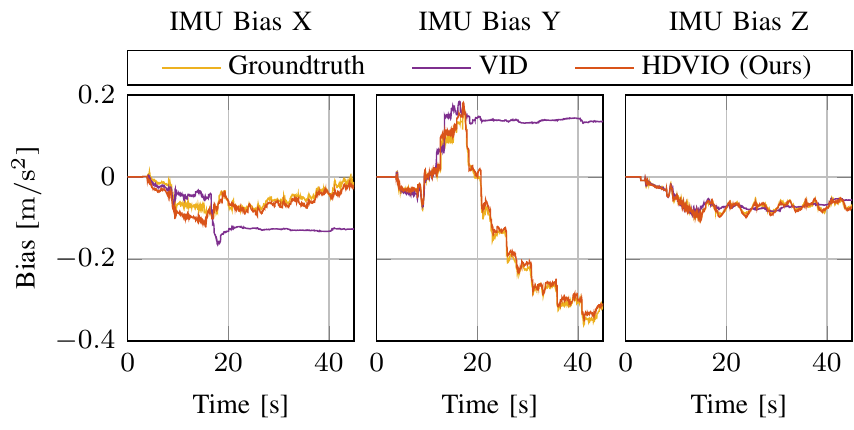}
    \vspace*{6pt}
    \caption{Accelerometer bias estimates from our real-world experiments. The ground-truth bias is obtained from the VIO system. The estimates of our \method\ match the ground-truth values, while, VIMO estimates diverge along the x-axis.}
    \label{fig:biases_realworld}
    \vspace*{-6pt}
\end{figure}

As stated by the authors~\cite{nisar2019vimo}, the measurement model in VIMO, in case of continuous external disturbances, introduces an inconsistency in the estimates of the accelerometer bias resulting in decreased accuracy of the motion estimate.
We show in Fig.~\ref{fig:biases_realworld} the accelerometer bias estimated by the VIO algorithm, by VIMO, and by our method. This is done for a sequence where the quadrotor, equipped with the drag board, flies a circle trajectory.
We do not have access to the ground-truth accelerometer bias. However, we consider the one estimated by the VIO algorithm as a good approximation of the ground-truth value since the VIO achieves very high performance in this sequence, cf. Table~\ref{tab:realworld_results}, thanks to the high number of visual features being tracked.
The bias estimate of \method\ closely follows the one of the VIO system, while the estimate of VIMO converges to a wrong value along the x-axis.
We include the position and orientation absolute trajectory errors in Table~\ref{tab:realworld_results}.
In all 4 sequences, the rich texture environment simplifies the pose estimation problem.
For this reason, the three algorithms achieve similar accuracy.

\section{Discussion}\label{sec:Discussion}
The proposed hybrid dynamics model, combining a point-mass quadrotor model with a learning-based residual force term, overcomes the limitations of the state-of-the-art visual-inertial-model-based odometry system, VIMO, in the case of large model mismatch (high speeds, systematic noise) and continuous external disturbances (continuous wind).
Indeed, our \method\ increases the accuracy of both motion and external force estimation.

Our learning-based module outperforms first-principle state-of-the-art quadrotor models, which have access to the full state of the quadrotor, in predicting the aerodynamic drag force.
In contrast to these methods, our \method\ only relies on thrust and gyroscope measurements, which are commonly available on any commercial drone platform. 

Another advantage of our approach is that our network does not rely on ground-truth thrust forces to supervise training.
In fact, our training strategy consists of minimizing the difference between
the relative position and velocity changes predicted by propagating our hybrid drone dynamics with respect to the supervision quantities.
Obtaining the supervision signal does not require access to expensive motion-capture systems but can be obtained from SLAM solutions~\cite{cioffi2022continuous, schoenberger2016sfm}.
These SLAM solutions rely only on camera and IMU data, simplifying the training data collection process. 
In Fig.~\ref{fig:colmap}, we show such an experiment, where we trained the learning-based component purely with position and velocity signal obtained via SLAM. 
Furthermore, even a human pilot can be used to control the drone, as expert pilots typically control a drone by sending a collective thrust command along with the desired bodyrates~\cite{pfeiffer2022visual}. 
This simple training data collection alleviates a limitation of our hybrid drone model: while a single dynamics model works only for a specific drone, it is easy to record data for training a new model.

Our hybrid drone model exhibits strong generalization capabilities to velocities and trajectories unseen during training.
\camera{To show generalization to unseen velocities, we train \method$^*$ in Sec.~\ref{sec:BlackbirdDataset}, on a dataset containing speeds only up to 2 m/s (\method\ is trained with speeds up to 9 m/s). When tested on the full range of speeds (up to 8 m/s), we observe that \method$^*$ achieves a decrease in performance of only 4\% in $\text{ATE}_{\text{T}}$ and still outperforms the baselines (see Table~\ref{tab:blackbird}).
We show the generalization performance w.r.t. the kind of trajectory in different examples. In the NeuroBEM dataset (see Table~\ref{tab:forces_torques}), the test dataset contains 30\% of trajectories completely unseen during training, and 70\% is at least different in speed and size.
Our system outperforms BEM and PolyFit by 50\% and 20\%, respectively, and is only inferior to NeuroBEM which has access to the full vehicle state.
Furthermore, our method can predict the external force acting on the drone flying an unseen random trajectory 40\% more accurately than VIMO (see Fig.~\ref{fig:viddataset}). In Sec.~\ref{sec:RealworldExp}, we train our network exclusively on random trajectories and observe more accurate estimate of the wind force (see Fig.~\ref{fig:forces_realworld}) and accelerometer bias (see Fig.~\ref{fig:biases_realworld}).}

\camera{
Furthermore, \method\ achieves high robustness w.r.t. VIO failures and continuous external disturbances.
In Sec.~\ref{sec:BlackbirdDataset}, \method\ achieves the largest improvements, equal to 41\% and 33\%, on the fastest trajectory, Egg 8 m/s (see Table~\ref{tab:blackbird} and Fig.~\ref{fig:blackbird}).
Due to motion blur and fast yaw changes tracking features is difficult here, resulting in the VIO system accumulating large drift. Moreover, neglecting drag effects in the drone model, as in~\cite{nisar2019vimo, ding2021vid}, is not a proper assumption at this speed. 
We evaluate the ability to estimate external forces in the presence of continuous perturbations: pulling rope (see Sec.~\ref{sec:VIDDataset}) and wind (see Sec.~\ref{sec:Setup}). In all these challenging scenarios, \method\ outperforms the baselines, by up to 40\% in force prediction (see Fig.~\ref{fig:viddataset}), highlighting its robustness.}

\camera{In this work, similarly to VIMO, the model is assumed to be fixed during a flight. 
Changes in the mass (external payload) or in the actuation inputs (hardware degradation) are seen as an external force.
An interesting venue for future work is to train the neural network to estimate these model changes as residual forces.}

We decided to use in \method\, as well as in VIMO and in the VIO system without the dynamics model, the visual frontend proposed in~\cite{Forster17troSVO}.
Our choice is based on the high robustness achieved by~\cite{Forster17troSVO} thanks to its semi-direct approach to visual feature tracking and low computational requirements. 
These characteristics are very appealing for VIO applications onboard flying vehicles.
\section{Conclusions}\label{sec:Conclusions}
 This work proposes a novel method to model the quadrotor dynamics in visual-inertial odometry systems.
Our dynamics model combines a first principles quadrotor model with a learning-based component that captures unmodeled effects, such as aerodynamic drag.
The proposed method overcomes the limitations of the state-of-the-art visual-inertial-model-based odometry system, VIMO, by increasing the accuracy of motion and external force estimation up to $33\%$ and $40\%$, respectively.
Our learning-based component shows strong generalization capabilities beyond the type and speed of the trajectories seen in the training dataset.
An evaluation of the accuracy of the residual force estimates shows that our learning-based component outperforms sophisticated first-principle models that have access to the full state of the quadrotor.
Experiments in controlled wind conditions show that our hybrid dynamics model achieves accurate predictions of the force affecting the quadrotor due to continuous wind.

Our \method\ method can increase the safety of autonomous flights in hazardous scenarios, such as fast flights and operations in windy conditions. In view of the increasing drone usage in our everyday life, such aspects are becoming more and more relevant and we believe that our work makes a valuable contribution towards this goal.

{
\bibliographystyle{unsrtnat}
\bibliography{references}}
\clearpage

\section{Appendix}\label{sec:Appendix}

\subsection{Comparison against VID-Fusion}

VID-Fusion is a visual-inertial-model-based odometry method very similar to VIMO.
The main difference between the two methods is the external force model.
While in VIMO, the external force is modeled as a Gaussian random variable with zero mean, in VID-Fusion, the mean is obtained by integrating the difference between accelerometer and thrust measurements.
According to the authors, this prior helps the estimation of continuous external forces.
In the main paper, we did not include VID-Fusion~\cite{ding2021vid} among the baselines for conciseness. Since the drone model is the same as the one used in VIMO, VID-Fusion has the same limitations. 
Consequently, including VID-Fusion in the baselines does not affect the conclusions drawn in the paper. 

In this appendix, for the sake of completeness, we present the same experiments as in the main paper including VID-Fusion among the baselines. 
We refer the reader to the main paper for the detailed description of the experimental setup and focus here only on the results.\newline

\subsubsection{Blackbird Dataset}

We present the $\text{ATE}_{\text{T}}$ and $\text{ATE}_{\text{R}}$ on the evaluation sequences of the Blackbird dataset in Table~\ref{tab:blackbird_appendix}.
Since VIMO and VID-Fusion use the same drone dynamics model and there are no external perturbances acting on the drone in this dataset, their trajectory estimation accuracy is similar.
Our method outperforms both baselines as well as the VIO solution.
Large improvements are in the fast sequences, where our learned aerodynamics model brings additional information to the VIO backend.
However, the performance difference is small in slow sequences, where including aerodynamic effects in the drone model is less effective.
\begin{table}[h!]
\caption{
Evaluation of the trajectory estimates in the Blackbird dataset. In bold are the best values, and the second-best values are underlined.}
\vspace{3pt}
\label{tab:blackbird_appendix}
\setlength{\tabcolsep}{4pt}
\begin{tabularx}{1.0\linewidth}{P{1.2cm}P{0.8cm}|C|C|C|C}
\toprule 
\multirow[c]{2}{=}[-6pt]{\centering Trajectory\newline Name} & \multirow[c]{2}{=}[-6pt]{\centering $v_{\max}$ \newline [\unit{m/s}]} & \multicolumn{4}{c}{Evaluation Metric: $\text{ATE}_\text{T}$ [\unit{m}] / $\text{ATE}_\text{R}$ [\unit{deg}]}  \\[4pt]
& & VIO & VID & VIMO & \method \newline (ours) \\
 \midrule
 Bent Dice & 3 & \textbf{0.20} / 1.78 & 0.25 / \textbf{1.18} &  0.31 / \underline{1.53} & \underline{0.21} / \underline{1.53} \\
 Clover & 5 & 0.90 / 3.52 & \underline{0.83} / \underline{2.48} & 0.88 / 3.66 & \textbf{0.60} / \textbf{2.08}\\
 Egg & 5 & 1.07 / 1.54 & 0.81 / 1.61 & \underline{0.75} / \underline{1.34} & \textbf{0.59} / \textbf{1.21} \\
 Egg & 6 & 1.40 / \underline{2.35} & 1.10 / 2.42 & \underline{0.98} / 4.89 & \textbf{0.83} / \textbf{1.62} \\
 Egg & 8 & 1.79 / 4.55 & \underline{1.47} / 4.84 & 1.57 / \underline{3.69} & \textbf{1.06} / \textbf{2.89} \\
 Mouse & 5 & 1.10 / 4.54 & \underline{0.54} / \underline{2.10} & 0.76 / 2.14 & \textbf{0.36} / \textbf{1.40} \\
 Star & 1 & \underline{0.17} / 0.78 & 0.18 / \textbf{0.54} & 0.18 / 1.05 & \textbf{0.16} / \underline{0.58} \\
 Star & 3 & 0.62 / 3.50 & 0.50 / 2.93 & \underline{0.43} / \textbf{1.38} & \textbf{0.38} / \underline{1.40} \\
 Winter & 4 & 0.97 / 2.92 & \underline{0.66} / \underline{2.05} & 0.69 / 2.46 & \textbf{0.57} / \textbf{1.54} \\
\bottomrule
\end{tabularx}
\end{table}

\subsubsection{VID Dataset}

\begin{figure}[]
    \centering
    \includegraphics{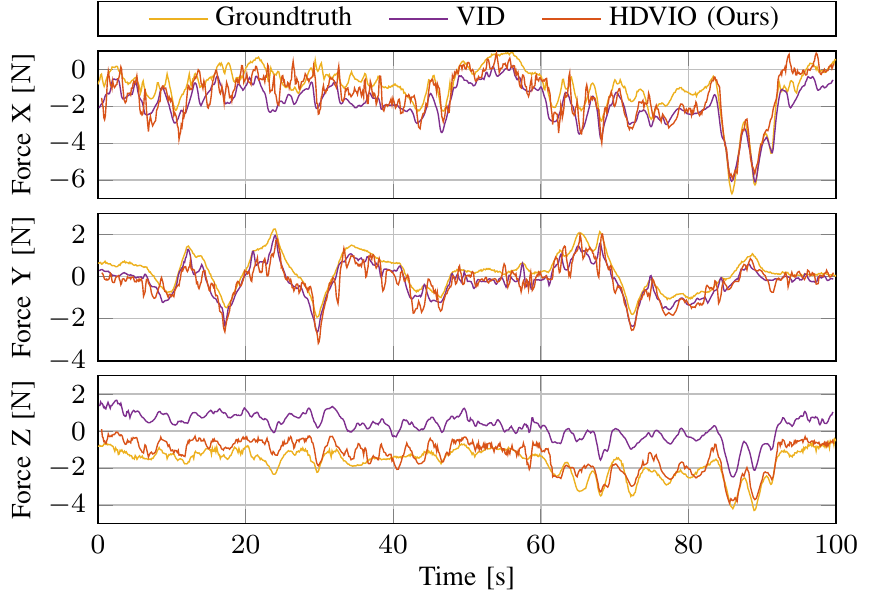}
    \caption{Comparison of the external force estimate in \textit{the sequence 17} of the VID dataset. Comparison of the external force estimate. \method\ drastically improves the force estimation along the z-axis resulting in a $42\%$ reduction of the RMSE compared to VID-Fusion.}
    \label{fig:viddataset_appendix}
    \vspace{-9pt}
\end{figure}

\begin{figure}[t]
    \centering
    \includegraphics{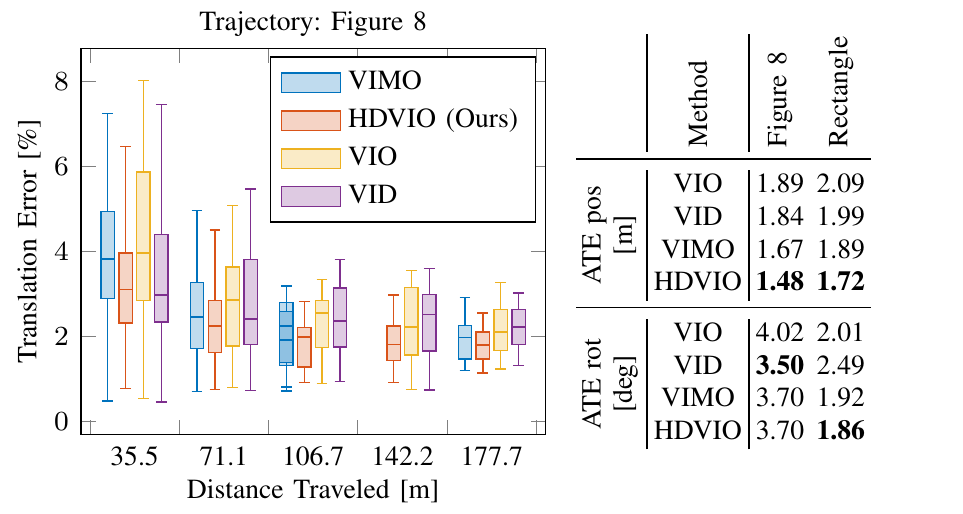}
    \caption{The plot and accompanying table show how our \method\ performs in a setting where the training data for the learning-based component is gathered from a vision-based SLAM system in the outdoor sequences of the VID dataset. The flown trajectories are at low speeds below \unit[3]{m/s}, which is why all four methods show good performance, with \method\ being the more accurate.}
    \label{fig:colmap_appendix}
    \vspace{-15pt}
\end{figure}

We evaluate the estimates of the external force in the \textit{sequence 17} of the VID dataset in Fig~\ref{fig:viddataset_appendix}.
In this sequence, the quadrotor is attached to a rope.
Ground-truth forces are available from a force sensor attached to the other end of the rope.
The force estimates are aligned to the motion-capture reference frame using the \textit{posyaw} alignment method~\cite{Zhang18iros}.
Our hybrid drone model has learned to compensate for a systematic residual error affecting the thrust inputs. 
We believe that the cause of this error
is inaccuracy in the rotor/thrust coefficients used to compute the
collective thrust inputs from the rotor speed measurements.
It is visible from the force estimates along the z-axis, that, VID-Fusion, similar to VIMO (see Fig.~\ref{fig:viddataset}), is not able to compensate for this systematic error.
The RMSE achieved by VID-Fusion along the z-axis is~\unit[1.95]{N}. The overall RMSE achieved by VID-Fusion is~\unit[1.12]{N}.
The RMSE achieved by \method\ along the z-axis is~\unit[0.55]{N}. The overall RMSE achieve by \method\ is \unit[0.65]{N}.
We do not include the estimates of VIMO in Fig.~\ref{fig:viddataset_appendix} for the sake of readability.
The $\text{ATE}_{\text{T}}$ achieved by VID-Fusion is the same as the one achieved by VIO, VIMO, and, \method\, namely \unit[0.02]{m}.

For completeness, we evaluate VID-Fusion also on the outdoor sequences.
We present the relative errors and the $\text{ATE}_{\text{T}}$ and $\text{ATE}_{\text{R}}$ in Fig.~\ref{fig:colmap_appendix}.
In these experiments, our hybrid drone model has been trained without using an external motion-capture system.
The flown trajectories are at low speeds below 3 m/s, which is why
all four methods show good performance, with \method\ being
the more accurate.
\begin{figure*}[t!]
    \centering
    \includegraphics{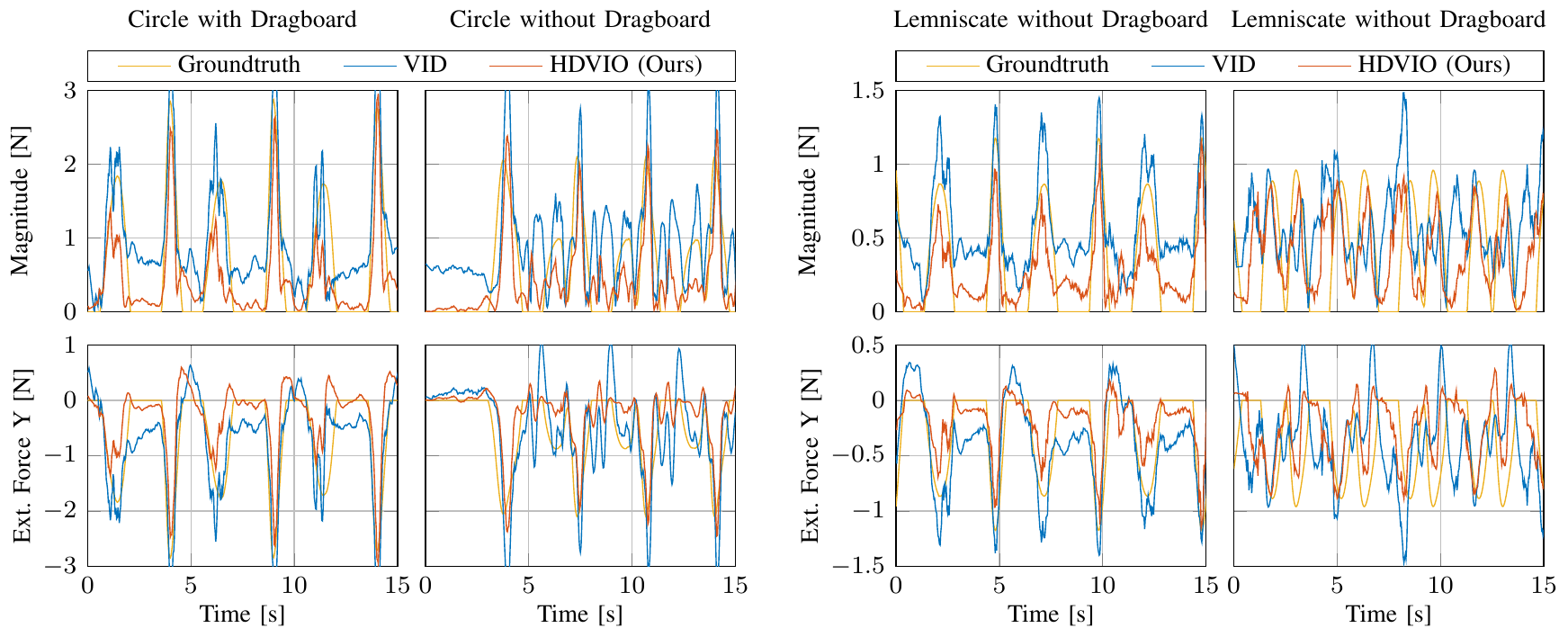}
    \caption{Wind disturbance estimates in our real-world experiments. The magnitude and the y-axis component of the wind force estimated by \method\ and VID-Fusion. Left: circle trajectory. Right: Lemniscate trajectory. In all the plots, it is visible that \method\ achieves more accurate force estimates than VID-Fusion.}
    \label{fig:forces_realworld_appendix}
    \vspace*{-9pt}
\end{figure*}

\subsubsection{Real-world Experiments}
\begin{figure}[t!]
    \centering
    \includegraphics{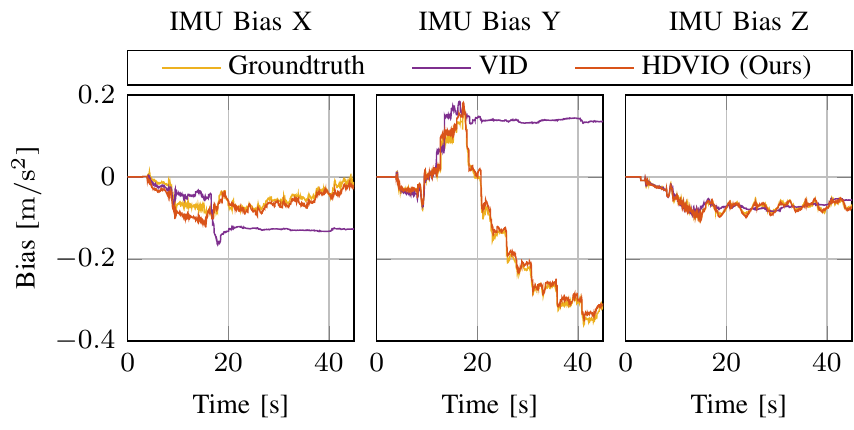}
    \vspace*{6pt}
    \caption{Accelerometer bias estimates from our real-world experiments. The ground-truth bias is obtained from the VIO system. The estimates of our \method\ match the ground-truth values, while, VID-Fusion estimates diverge along the x-axis.}
    \label{fig:biases_realworld_appendix}
    \vspace*{-6pt}
\end{figure}

The estimates of the external force of VID-Fusion due to the wind gusts are in Fig.~\ref{fig:forces_realworld_appendix}. 
We show the force along the y-axis, which is the direction of the wind gusts, after alignment to the world frame using the ground-truth orientations.
From Fig.~\ref{fig:forces_realworld_appendix}, it is clear that \method\ outperforms VID-Fusion.
Notably, VID-Fusion overestimates the wind force when the quadrotor enters the wind field.
The reason for this behavior is that VID-Fusion jointly estimates the drag force with the external force.

Similar to VIMO, the measurement model in
VID-Fusion introduces inconsistency in the estimates of the accelerometer bias resulting in decreased accuracy of the motion
estimate.
We show in Fig.~\ref{fig:biases_realworld_appendix} the accelerometer bias estimated by the VIO algorithm, by VID-Fusion, and by our method. 
This is done for a sequence where the quadrotor, equipped with the
dragboard, flies a circle trajectory. 
Here, we consider the bias estimated by the VIO algorithm as a good approximation of the ground-truth values since the VIO achieves very high performance in this sequence thanks to the large number of visual features being tracked. 
The bias estimate of \method\ closely matches the one of the VIO system, while the estimate of VID-Fusion converges to wrong values. 
We include the position and orientation absolute
trajectory errors in Tab.~\ref{tab:realworld_results_appendix}. In all 4 sequences, the rich texture environment simplifies the pose estimation problem. For this reason, the four algorithms achieve similar accuracy.

\begin{table}[t!]
    \centering
    \setlength{\tabcolsep}{3pt}
    \caption{Experimental results from our real-world experiments. We use (d) to indicate that a dragboard was attached to the drone.}
    \begin{tabularx}{1\linewidth}{l|CCCc|CCCc}
    \toprule
         &  \multicolumn{4}{c|}{ATE Position [m]}   & \multicolumn{4}{c}{ATE Rotation [deg]} \\
         & VIO & VID & VIMO & \method & VIO & VID & VIMO & \method \\[3pt]
         \midrule
         Circle (d) & \bf 0.07 & 0.10 & 0.1 & \bf 0.07 & 2.02 & 2.31 & \bf 1.80 & 2.06 \\
         Circle & \bf 0.06 & \bf 0.06 & 0.08 & \bf 0.06 & 1.21 & 1.37 & 1.19 & \bf 1.17 \\
         Lemniscate (d) & 0.38 & 0.53 & 0.34 & \bf 0.30 & \bf 2.39 & \bf 2.39 & 2.93 & 2.81 \\
         Lemniscate & 0.27 & 0.28 & 0.32 & \bf 0.20 & 2.44 & 2.05 & 1.93 & \bf 1.84 \\
         \bottomrule
    \end{tabularx}
    \label{tab:realworld_results_appendix}
\end{table}

\subsection{Clarification on Training with Vision-based SLAM Supervision}

We show in Fig.~\ref{fig:colmap}, that our hybrid drone model can be trained using pose supervision from a vision-based SLAM system~\cite{cioffi2022continuous} on the VID dataset.
However, the improvement in pose estimates is small compared to the baselines because these trajectories contain slow flights.

An alternative would have been to train with supervision from the vision-based SLAM system on the Blackbird dataset which contains faster flights.
However, this was not possible because most of the train sequences did not include images at the time this work was carried out.

\end{document}